\documentclass[letterpaper,10pt,journal,twoside]{IEEEtran}

\IEEEoverridecommandlockouts                              

\usepackage{graphicx,amsmath,amssymb,bm,xcolor,soul,nicefrac,listings,dsfont,caption,subcaption,wrapfig,sidecap} 
\usepackage[hidelinks]{hyperref}
\usepackage{courier}
\usepackage[algo2e]{algorithm2e} 

\SetCommentSty{mycommfont}
\usepackage{algorithmic}
\usepackage{booktabs}
\usepackage{siunitx}
\usepackage{algorithm}

\newcommand{\trsp}{{\scriptscriptstyle\top}}

\newcommand{\cov}{\mathrm{cov}}

\newcommand{\new}{{\!\scriptscriptstyle\mathrm{new}}}

\newrobustcmd{\B}{\bfseries}

\begin{document}

\title{Online Learning of Continuous Signed Distance Fields Using Piecewise Polynomials}

\author{Ante Mari\'c, Yiming Li, and Sylvain Calinon
\thanks{Manuscript received: December 21, 2023; Revised: March 17, 2024; Accepted: April 15, 2024.}%
\thanks{This paper was recommended for publication by Editor Jens Kober upon evaluation of the Associate Editor and Reviewers' comments.
This work was supported by the State Secretariat for Education, Research and Innovation in Switzerland and the European Commission’s Horizon Europe Program through the INTELLIMAN project (https://intelliman-project.eu/, HORIZON-CL4-Digital-Emerging Grant 101070136) and the SESTOSENSO project (http://sestosenso.eu/, HORIZON-CL4-Digital-Emerging Grant 101070310).}
\thanks{The authors are with the Idiap Research Institute, CH-1920 Martigny, Switzerland and also with the EPFL, 1015 Lausanne, Switzerland. 
            {\\\tt\footnotesize ante.maric@idiap.ch; yiming.li@idiap.ch; sylvain.calinon@idiap.ch}}%
\thanks{Digital Object Identifier (DOI): see top of this page.}
}

\markboth{IEEE Robotics and Automation Letters. Preprint Version. Accepted April, 2024}
{Mari\'c \MakeLowercase{\textit{et al.}}: Online Piecewise Polynomial SDF} 

\maketitle

\begin{abstract}
Reasoning about distance is indispensable for establishing or avoiding contact in manipulation tasks. To this end, we present an online approach for learning implicit representations of signed distance using piecewise polynomial basis functions. Starting from an arbitrary prior shape, our method incrementally constructs a continuous and smooth distance representation from incoming surface points, with analytical access to gradient information. The underlying model does not store training data for prediction, and its performance can be balanced through interpretable hyperparameters such as polynomial degree and number of segments. We assess the accuracy of the incrementally learned model on a set of household objects and compare it to neural network and Gaussian process counterparts. The utility of intermediate results and analytical gradients is further demonstrated in a physical experiment. For code and video, see \url{https://sites.google.com/view/pp-sdf/}.
\end{abstract}


\section{Introduction}
\IEEEPARstart{S}{cene} representation is a naturally emerging topic in robotics as a basis for physical interaction. In recent years, implicit modeling methods have been used as compact representations of environment properties such as distance, occupancy, and color. Signed distance functions (SDFs) model distances to closest occupied points by assigning zero values to surfaces, negative values to surface interiors, and positive values elsewhere. Previously used in environment mapping and collision avoidance settings \cite{oleynikova2017voxblox}, they have recently seen use in robotic manipulation as the field moves towards exploring contact-rich behaviors \cite{li2023learning}. In such scenarios, distance representations can be exploited to quickly and robustly retrieve gradients for a variety of tasks.
Furthermore, modeling the full range of distances, as opposed to only the zero level set, can be beneficial for reasoning about making or breaking contact, deformation, or penetration, with extensions to active agents such as users
or robots.
\begin{figure}[h!]
    \centering
    \includegraphics[width=0.67\linewidth]{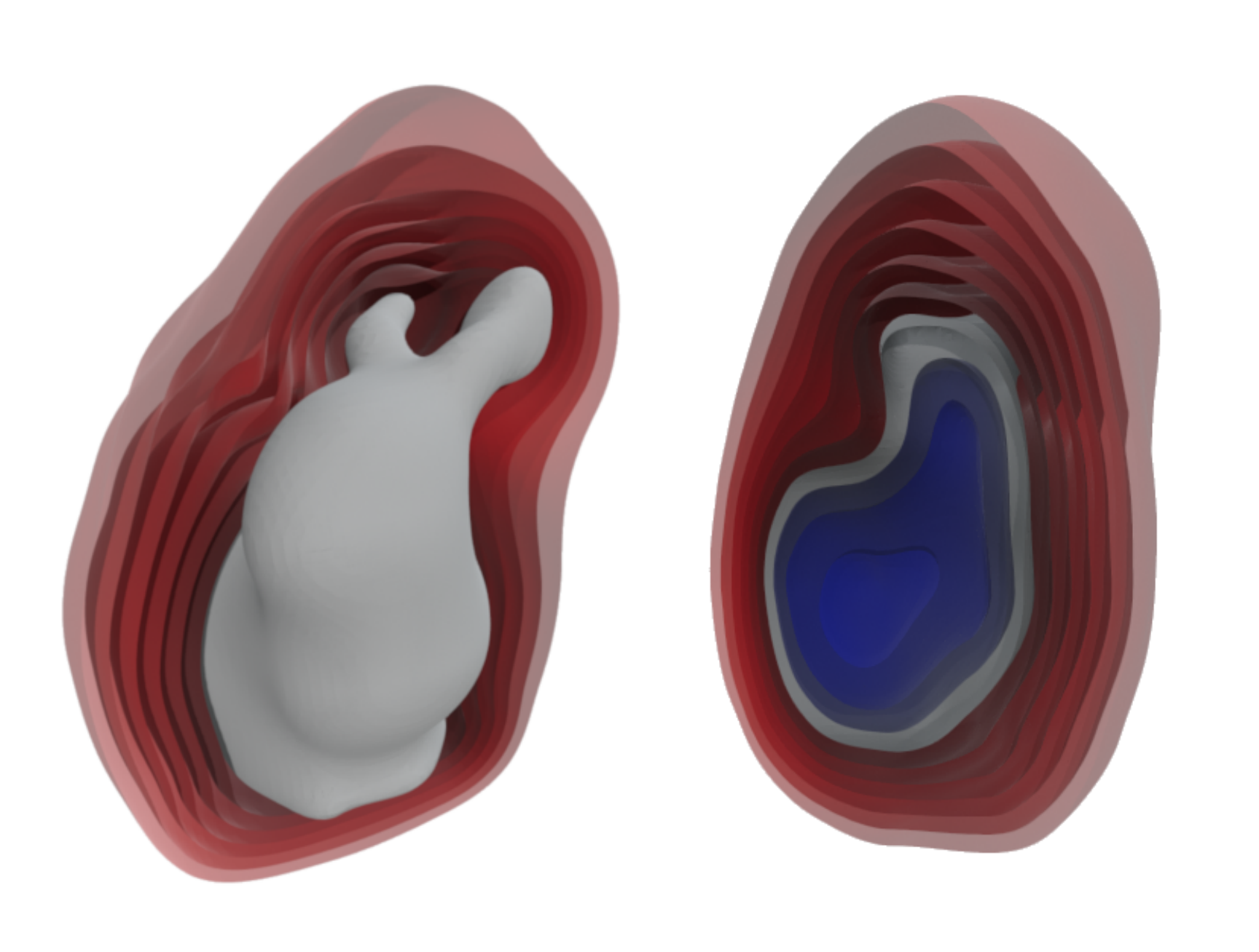}
    \caption{Mesh and distance level sets of the \textit{Stanford Bunny} reconstructed from piecewise polynomial basis functions. The model is learned incrementally from $1283$ %
    randomly sampled surface points and corresponding normal vectors.}
    \label{fig:bunny}
\end{figure}
Commonly used implicit SDF models in robotics rely on neural architectures
or Gaussian process models,
while alternative formulations remain less explored. Earlier computer graphics work encodes SDFs using basis functions, with recent extensions showing promising results using piecewise polynomial representations \cite{Pujol2023piecewise}. In robotics, basis functions have been used to encode movements as superpositions of primitives \cite{Calinon19MM}.
Basis function representations of distance can consequently be seen as a step toward combining movement with shape primitives of robot environments.

When operating in previously unseen environments, incrementally building a distance model from incoming data enables the integration of feedback for more robust and adaptive behavior.
We formulate an online method for learning signed distance fields represented as piecewise polynomial basis functions. Our method uses a simple incremental least squares approach and regularization scheme to approximate distance fields from incoming surface points. The resulting representation is $C^1$ continuous, with analytical access to gradients for downstream use.
To achieve fast update time on modest hardware, performance of the piecewise polynomial representation can be balanced through interpretable hyperparameters such as polynomial degree and number of segments. It is also easy to incorporate priors through basis function superposition weights. We evaluate the accuracy of our method on a set of diversely shaped household objects, showing comparable results to Gaussian process and neural network baselines, while relying on a lower number of parameters, and without the need to store training data for prediction.
The usability of intermediate results and analytical gradients is further demonstrated in a physical experiment where we use an evolving representation learned from noisy and partial point cloud data to survey and grasp an object of interest.

\section{Implicit environment representations}
In robotics, widely adopted scene representations use mapping approaches that rely on discretized occupancy grids and account for uncertainty \cite{hornung13auro}. Subsequent methods explicitly store distance information for real-time usage in dynamic settings by introducing distance fields \cite{oleynikova2017voxblox, Han2019FIESTAFI}. 
Following advancements in computer vision, recent work has been exploring implicit representations as scalable and compact alternatives for describing scenes without storing data in dense grids. 
Implicit representations of distance, volume, and color information have found application in robotics, with initial uses in navigation \cite{Adamkiewicz2021VisionOnlyRN} and mapping \cite{imapSucar:etal:ICCV2021}. Recent work uses implicit representations as visual encodings for grasping \cite{breyer2020volumetric}, whole-body manipulation \cite{li2023learning}, human-robot interaction \cite{liu2022}, planning  \cite{Driess2021LearningMA}, and control \cite{li20213d}.

\subsection{Implicit signed distance functions}
Implicit signed distance functions model geometry through continuous functions, thus decoupling memory from spatial resolution. Seminal work utilizes neural networks for shape representation, showing higher performance than point cloud, mesh, or grid-based counterparts \cite{Park_2019_CVPR}. Additional efforts have been put into investigating regularization and modeling methods that allow learning such representations in online scenarios using raw point cloud data \cite{pmlr-v119-gropp20a, Ortiz:etal:iSDF2022}. Recent methods introduce neural architectures that jointly represent SDF and color to track and reconstruct unknown objects \cite{wen2023bundlesdf}.
Similarly, Neural radiance fields (NeRFs) \cite{mildenhall2020nerf} jointly encode volume density and color. They have garnered much attention as environment representations, with recent extensions targeting real-time rendering \cite{kerbl3Dgaussians} and dynamic scenes \cite{pumarola2020dnerf}. However, NeRFs do not offer direct access to distance or derivative information, which can be beneficial for interpreting task execution. Furthermore, neural representations often require large amounts of data and do not translate readily to lower data regimes found in modalities such as tactile or proximity sensing.

\subsection{Gaussian process implicit surfaces}
Probabilistic models like Gaussian process implicit surfaces (GPIS) have been used to represent distances with account for uncertainty \cite{dragiev2011}. To extend the approach for mapping purposes, scaling issues of the Gaussian process have been addressed through the use of clustering and hierarchical models  \cite{blee-icra19>}. Subsequent combinations with implicit regularization methods enable accurate modeling of unsigned distance \cite{wu2021LogGPIS}. These methods were later combined to give a unified mapping, odometry, and planning framework \cite{Wu2023MOP}.

\subsection{Basis function representations}
Basis functions can describe complex representations as weighted superpositions of simple signals. In robotics, they are well-known as the underlying representation used to encode movement primitives \cite{ijspeert2013dmp, ParaschosDPN2013}. A detailed review can be found in \cite{Calinon19MM}. Their role in computer graphics extends to higher-dimensional input space to represent shapes using SDFs \cite{Juttler2002}. Learning such representations from point clouds and normals can be achieved simply by solving a linear system of equations using a least squares approach or iterative optimization procedures \cite{Taubin2012SSD}.
A recent example utilizes piecewise polynomials to approximate high-fidelity SDFs using an adaptive grid \cite{Pujol2023piecewise}.
The resulting SDF models offer analytical access to distance and gradients, and a desired order of continuity can be ensured by constraining the superposition weights. The following section describes an online formulation of piecewise polynomial SDF based on incremental learning, with the aim of adaptively guiding movement in manipulation tasks.

\section{Piecewise polynomial SDF}
\label{sec:Method}

\subsection{Bernstein polynomial basis functions}

\begin{figure*}[h!]
    \centering
    \includegraphics[width=0.96\textwidth]{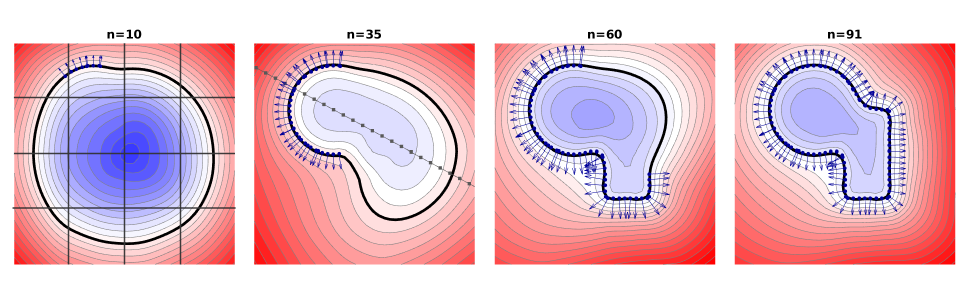}
    \caption{Incremental model updates used to model a 2D shape on a $4\times 4$ grid, starting from a circular prior. Sampled points and normals are shown in dark blue, and the reconstructed zero-level contour in black. The reconstructed SDF is visualized as a color map. The normal ray and regularization points of a single sample are displayed in the second image.
    }
    \label{fig:evolution}
\end{figure*}

The value of a univariate function $f(x)$ at input $x$ can be represented as a weighted sum of $K$ basis functions with
\begin{equation}
	f(x) = \sum_{k=0}^{K} {\phi}_{k}(x) \; w_k = \bm{\phi}(x) \; \bm{w},
\end{equation}
where $\bm{\phi}$ can come from any family of basis functions. For our SDF representation, we use Bernstein polynomial bases, which give smooth function approximations on bounded intervals. For degree $K$, they can be computed as
\begin{equation}
	{\phi}_{k}(x) = \frac{K!}{k!(K-k)!} \; {(1-x)}^{K-k}\; x^{k},
\end{equation}
$\forall k\in\{0,\ldots,K\}$.
Instead of considering a global encoding which might require the use of high-order polynomials, we split the problem into a set of local fitting problems that can consider lower-order polynomials.
We retain $C^1$ continuity by introducing constraints of the form
\begin{align}
\label{eq:constraint1}
    w_K^a &= w_0^b\\
    w_1^b &= -w_{K-1}^a+2w_0^b,
\label{eq:constraint2}
\end{align}
where $a$ and $b$ are concatenated polynomials of degree $K$, and $w_k^a$ is used to denote the $k$-th weight of polynomial $a$.
Polynomial bases and their derivatives can then be expressed in matrix form as
\begin{align}
\label{eq:matrix}
    \bm \phi (x) &= \bm T(x) \bm B \bm C,\\
    \frac{\partial\bm{\phi}(x)}{\partial x}&=\frac{\partial\bm{T}(x)}{\partial x} \bm{B}\bm{C},
\end{align}
with $\bm T (x) = \left[1~x~x^2~\cdots~x^K\right]$ being a polynomial feature map of input $x$, $\bm B$ the corresponding Bernstein coefficient matrix, and $\bm C$ a constraint matrix of the form
\begin{equation}
	\bm{C} = \begin{bmatrix}
	1 & 0 & \cdots & 0 & 0 & \cdots \\
	0 & 1 & \cdots & 0 & 0 & \cdots \\
	\vdots & \vdots & \ddots & \vdots & \vdots & \ddots \\
	0 & 0 & \cdots & 1 & 0 & \cdots \\
	0 & 0 & \cdots & 0 & 1 & \cdots \\
	0 & 0 & \cdots & 0 & 1 & \cdots \\
	0 & 0 & \cdots & -1 & 2 & \cdots \\
	\vdots & \vdots & \ddots & \vdots & \vdots & \ddots
	\end{bmatrix},
\end{equation}
enforcing \eqref{eq:constraint1} and \eqref{eq:constraint2}.

Successive Kronecker products can be used to extend the described representation to any number of input and output dimensions. For clarity and visualization purposes, we will continue the method description for a two-dimensional case. Namely, an extension to two-dimensional input space (Cartesian coordinates) and one-dimensional output (signed distance) can be calculated as
\begin{align}
    \label{eq:Psi2D_analytic}
	\bm{\Psi}(x, y) &= \bm{\phi}(x) \,\otimes\, \bm{\phi}(y),
\end{align}
with partial derivatives and gradient computed analytically as   
\begin{align}
	\frac{\partial\bm{\Psi}(x, y)}{\partial x} &= \frac{\partial\bm{\phi}(x)}{\partial x} \,\otimes\, \bm{\phi}(y),\\
	\frac{\partial\bm{\Psi}(x, y)}{\partial y} &= \bm{\phi}(x) \,\otimes\, \frac{\partial\bm{\phi}(y)}{\partial y},\\
    \bm{\nabla}\bm{\Psi}(x, y) &= \frac{\partial\bm{\Psi}(x, y)}{\partial x} \,\otimes\, \frac{\partial\bm{\Psi}(x, y)}{\partial y},
	\label{eq:parialPsi}
\end{align}

This can then be used to compute the distance and gradient values of the SDF at coordinate $(x, y)$ with
\begin{align}
    f(x, y) &= \bm{\Psi}(x, y) \, \bm{w},\\
	\bm{\nabla}f(x, y) &= \bm{\nabla}\bm{\Psi}(x, y) \, \bm{w}.
\end{align}
The same procedure can be applied to calculate higher-order derivatives for regularization purposes
\begin{align}
   \frac{\partial \bm\nabla f(x, y)}{\partial x} &= \frac{\partial\bm\nabla\bm \Psi(x, y)}{\partial x}\bm{w}.
\end{align}
\noindent
The above representation extends analogously to accommodate three-dimensional Cartesian coordinates as input by applying an additional Kronecker product in \eqref{eq:Psi2D_analytic}.

\subsection{Computation of weights}
\label{sec:Learning}
To approximate the SDF using polynomial basis functions, any method capable of solving a system of linear equations of the form $\bm A\bm w=\bm s$ can be employed. The simplest case can utilize a batch least squares estimate of the form
\begin{align}
	\bm{w} &= (\bm{A}^\top \bm{A})^{-1}\bm{A}^\top \bm{s},
    \label{eq:LSQ}
\end{align}
or ridge regression as the regularized variant \cite{ridgeReg}.

\begin{figure*}[h!]
    \centering
    \begin{subfigure}[t]{0.158\textwidth}
        \includegraphics[width=\textwidth]{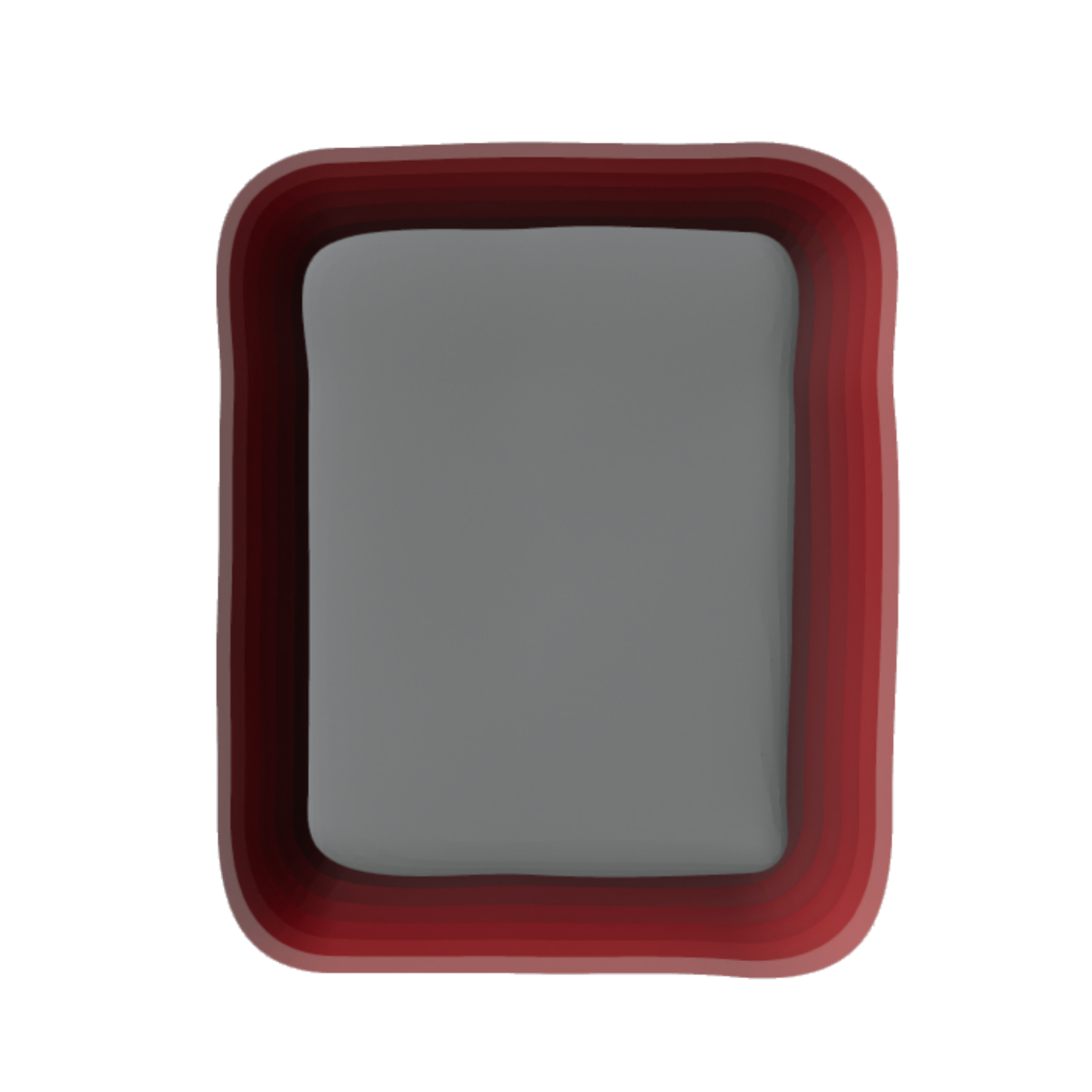}
        \caption{}
        \label{sub:cracker}
    \end{subfigure}%
    \begin{subfigure}[t]{0.158\textwidth}
        \includegraphics[width=\textwidth]{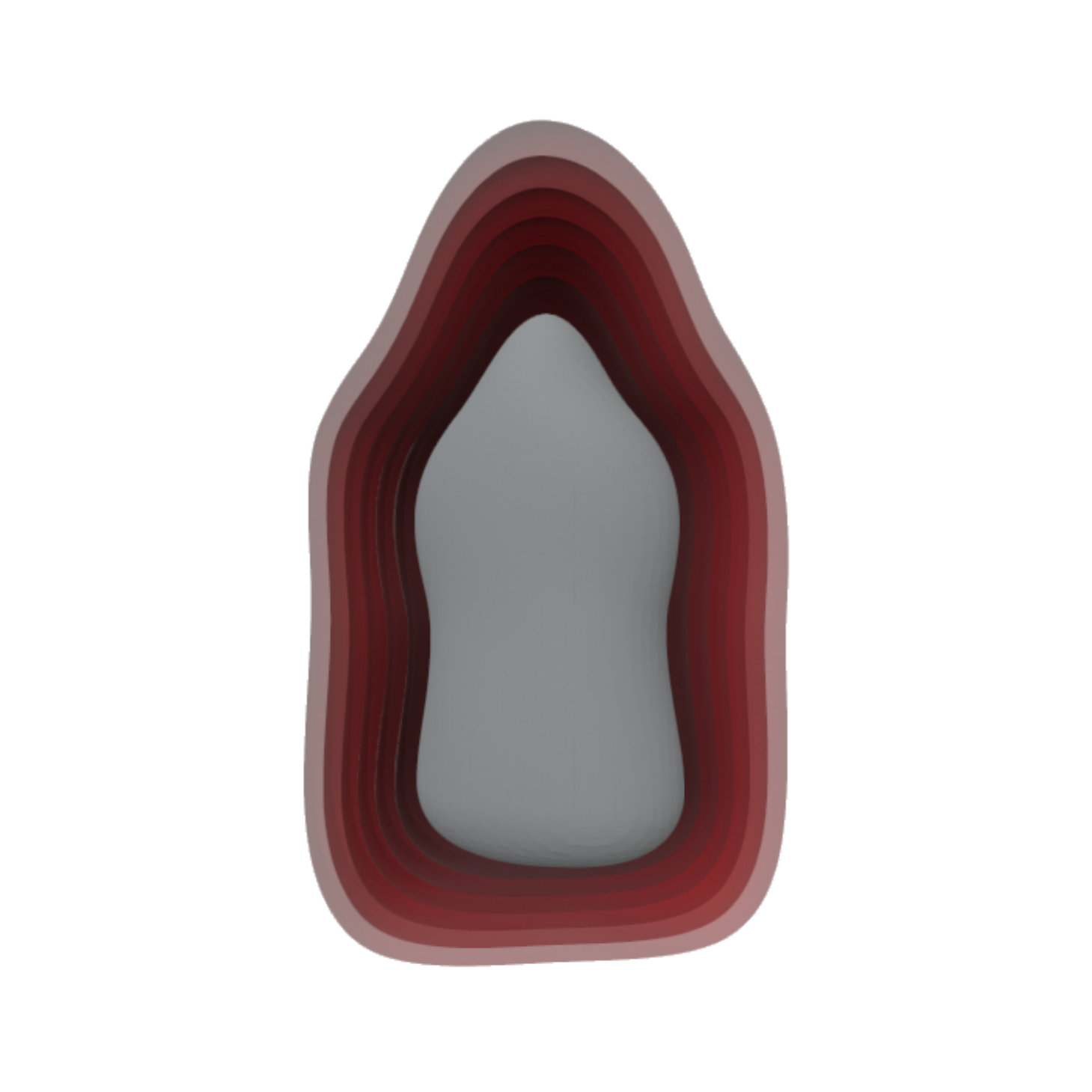}
        \caption{}
        \label{sub:mustard}
    \end{subfigure}
    \begin{subfigure}[t]{0.158\textwidth}
        \includegraphics[width=\textwidth]{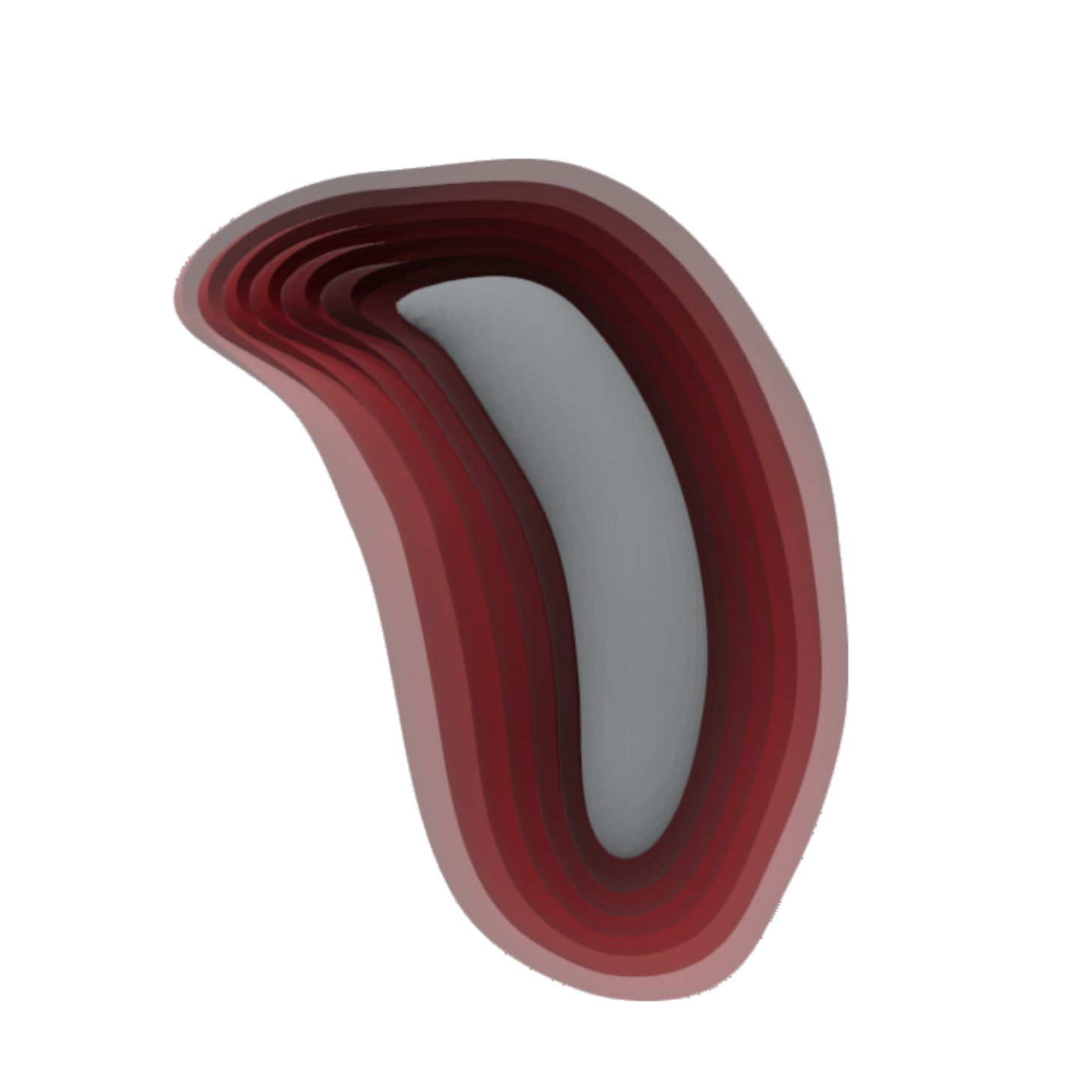}
        \caption{}
    \end{subfigure}
    \begin{subfigure}[t]{0.158\textwidth}
        \includegraphics[width=\textwidth]{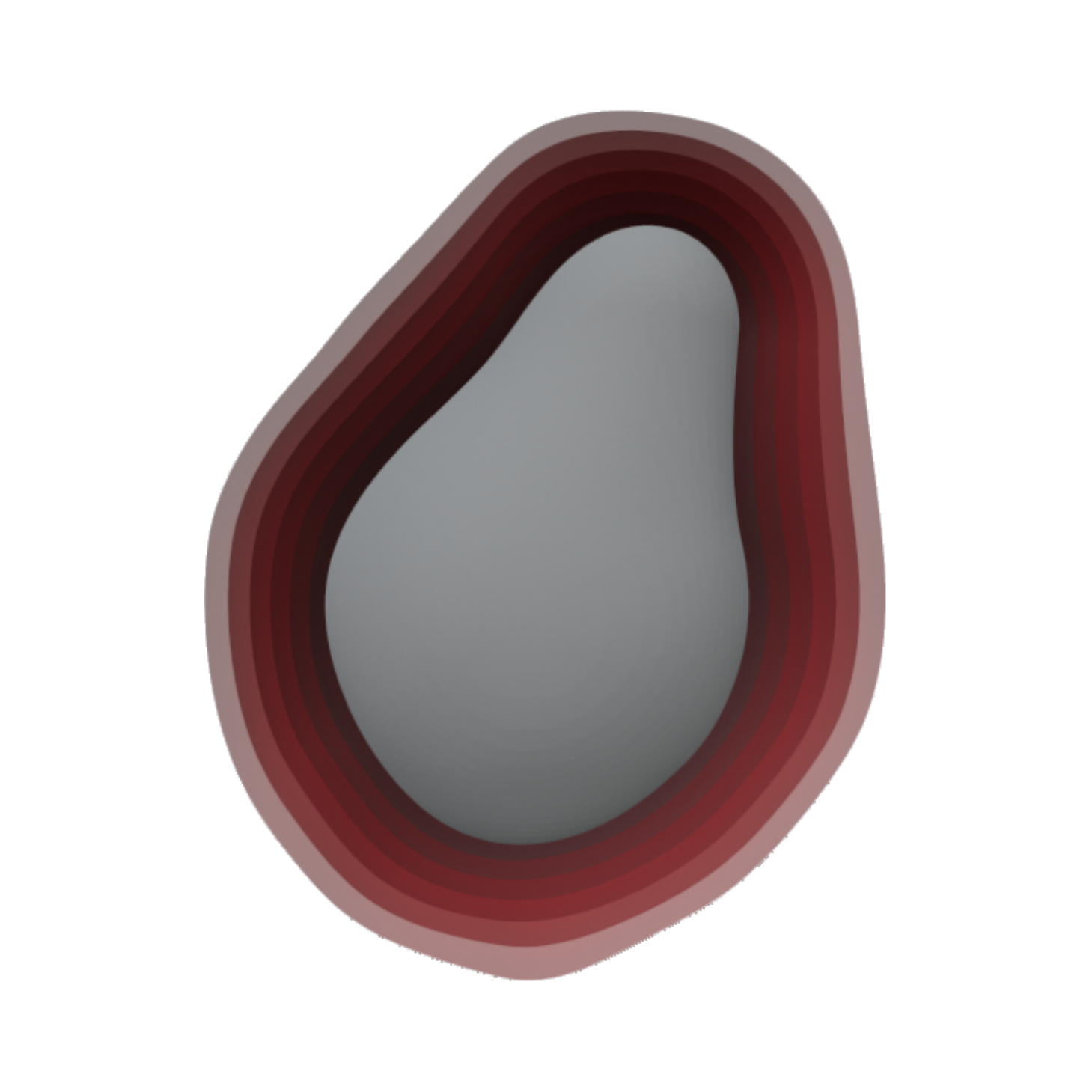}
        \caption{}
    \end{subfigure}
    \begin{subfigure}[t]{0.158\textwidth}
        \includegraphics[width=\textwidth]{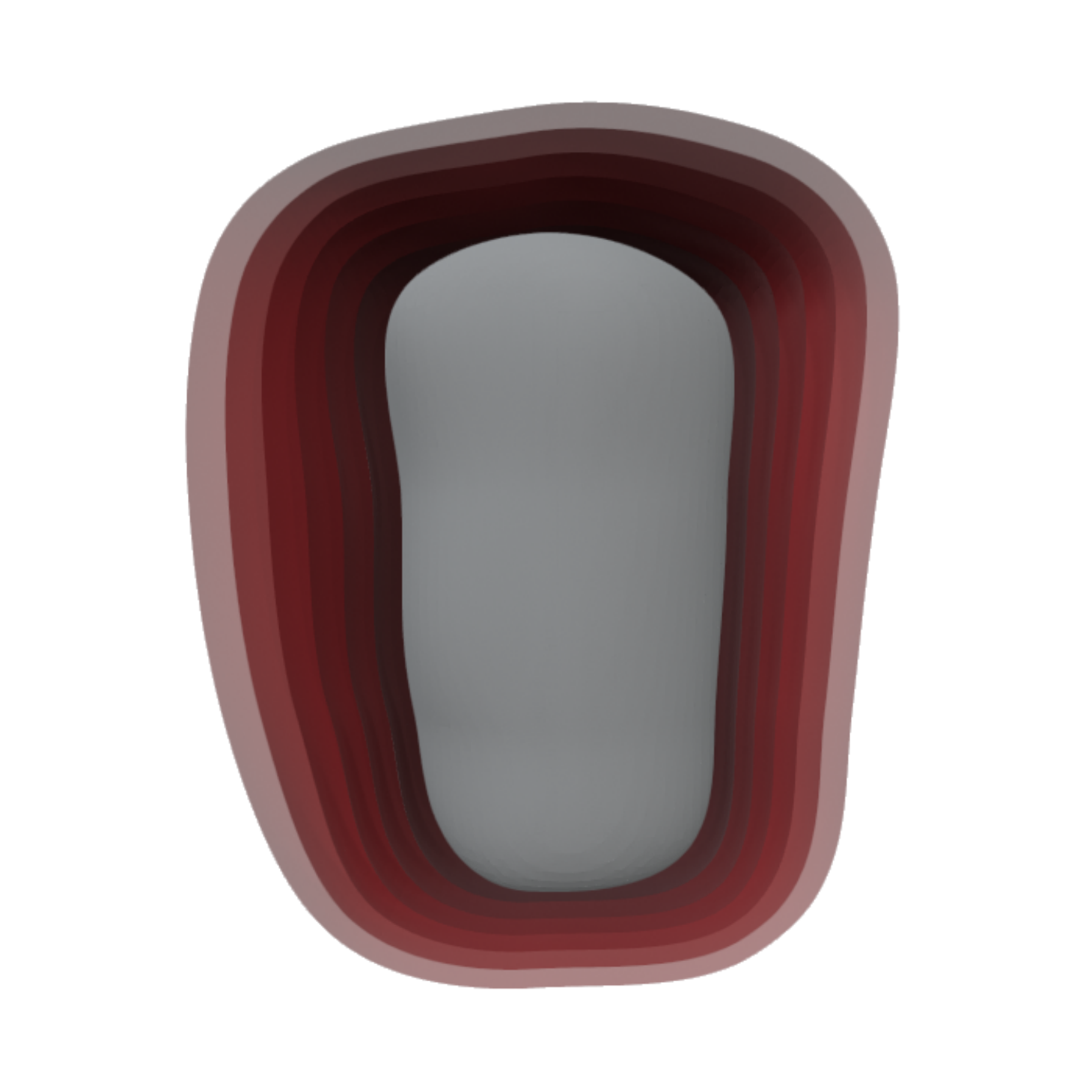}
        \caption{}
    \end{subfigure}\\
    \begin{subfigure}[t]{0.158\textwidth}
        \includegraphics[width=\textwidth]{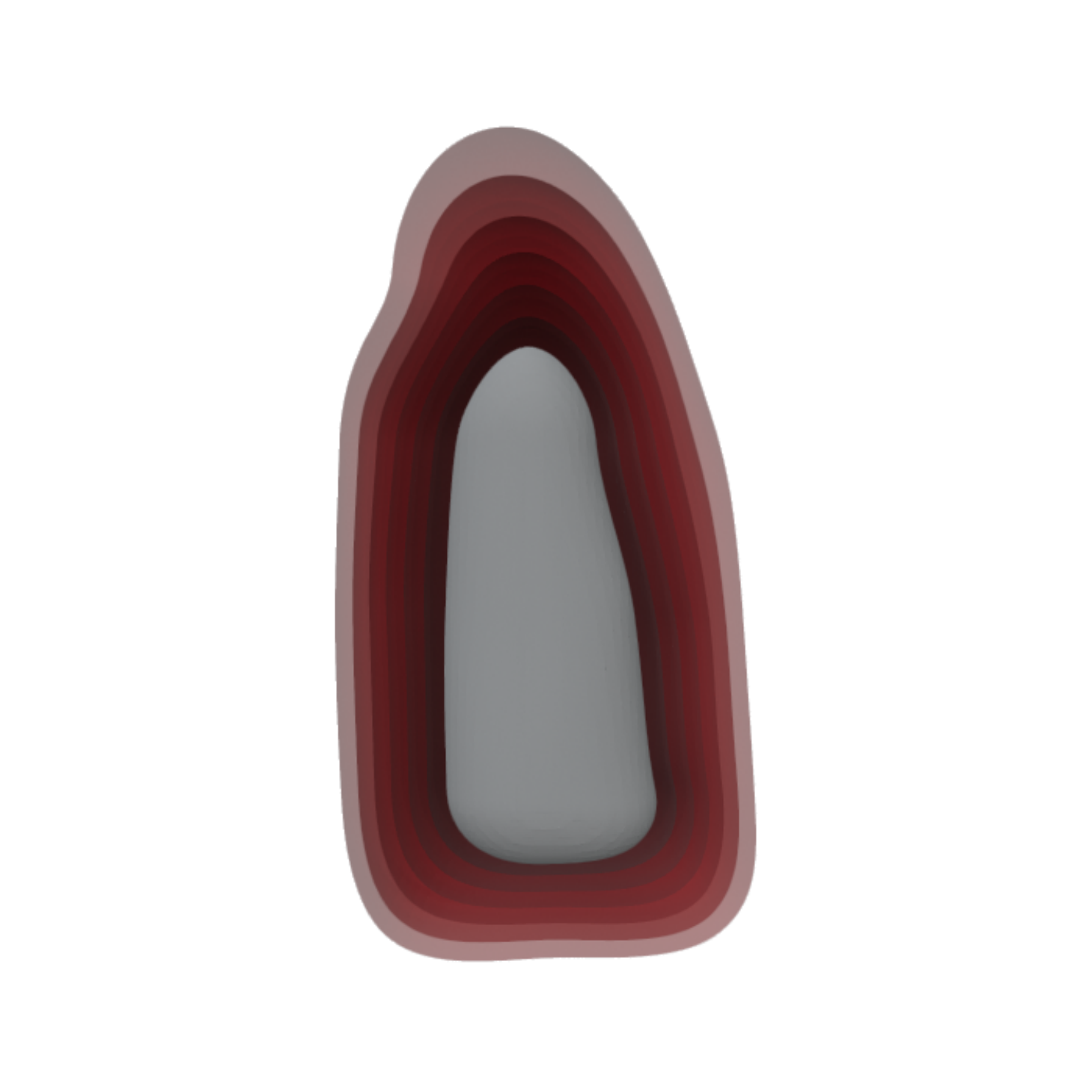}
        \caption{}
    \end{subfigure}
    \begin{subfigure}[t]{0.158\textwidth}
        \includegraphics[width=\textwidth]{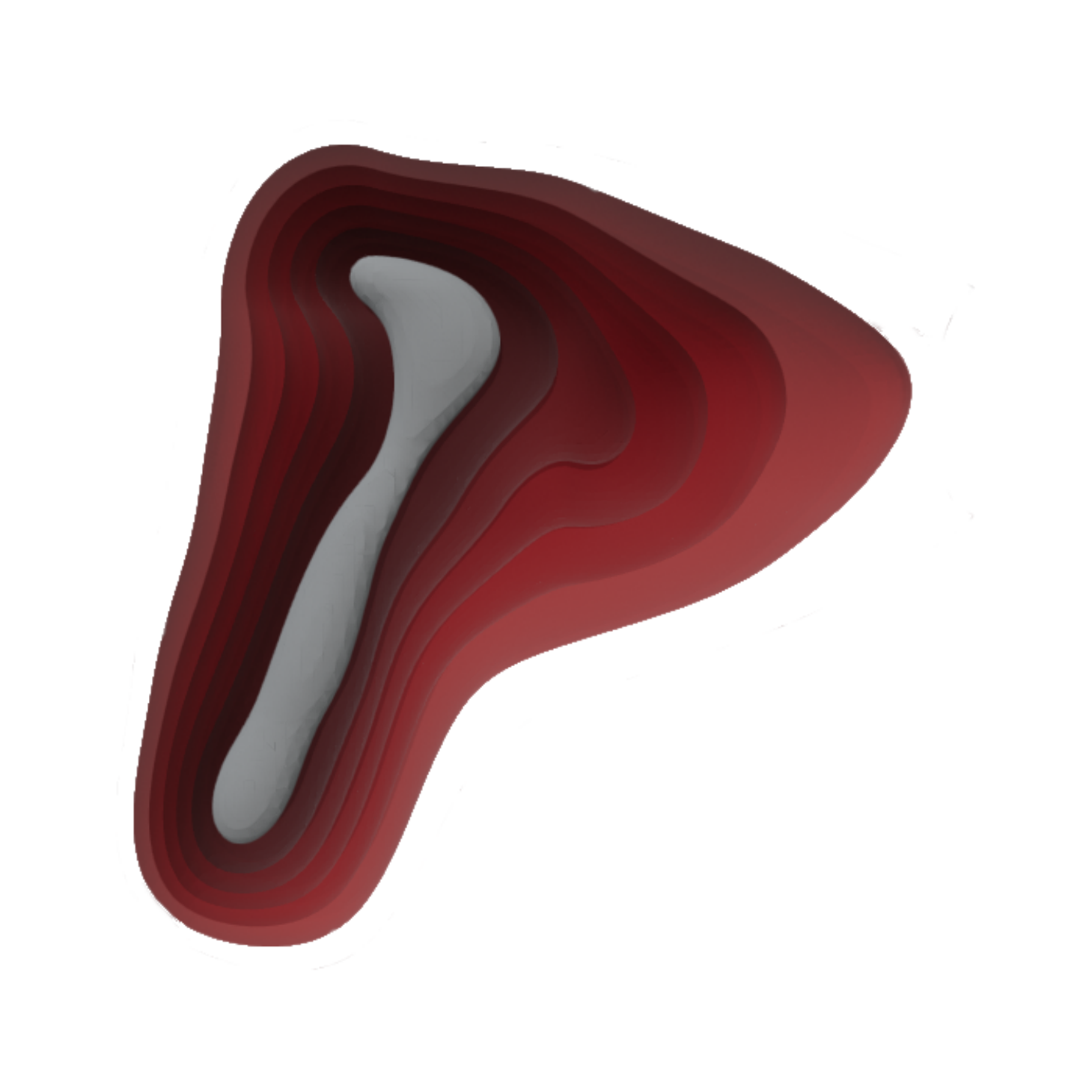}
        \caption{}
        \label{sub:hammer}
    \end{subfigure}
    \begin{subfigure}[t]{0.158\textwidth}
        \includegraphics[width=\textwidth]{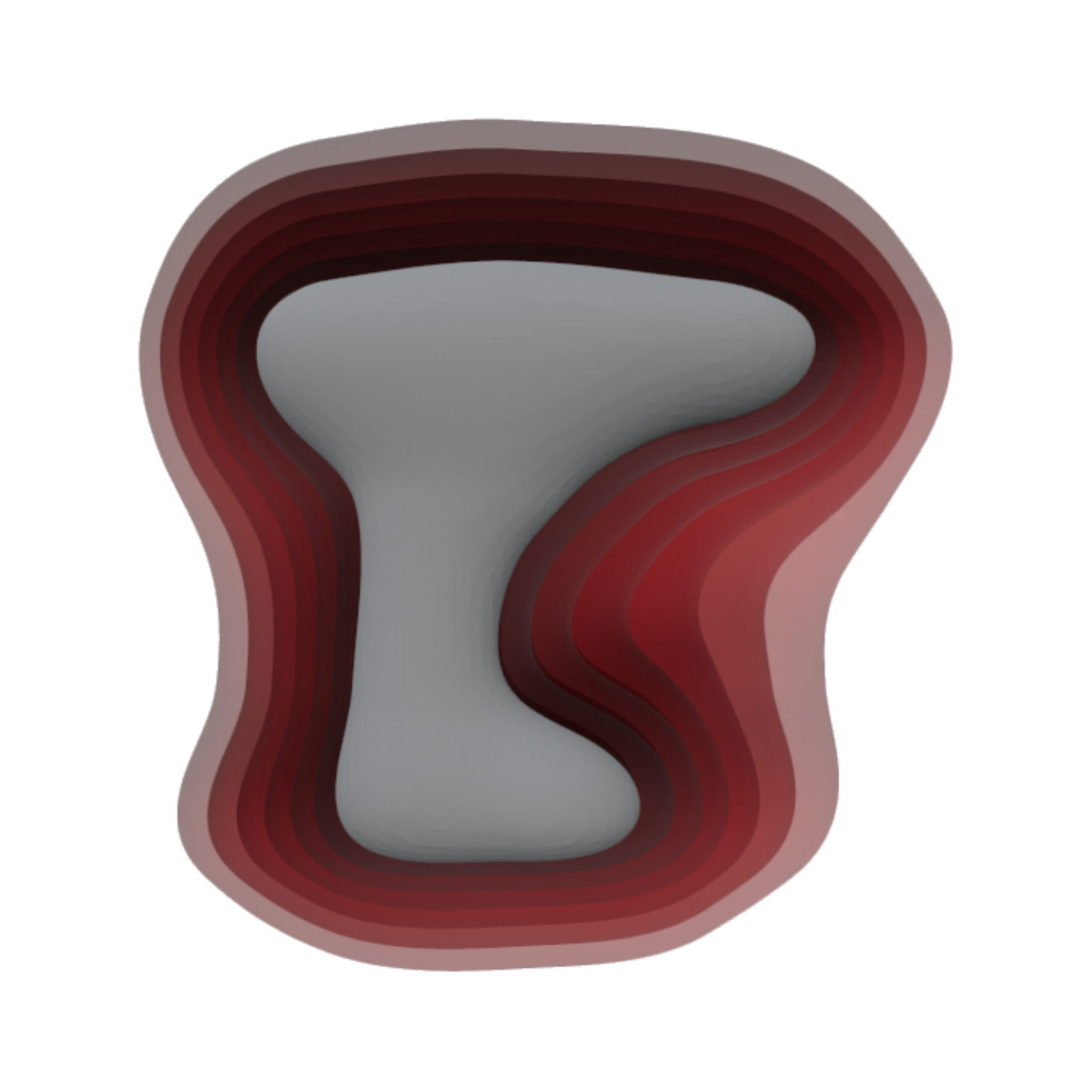}
        \caption{}
    \end{subfigure}
    \begin{subfigure}[t]{0.158\textwidth}
        \includegraphics[width=\textwidth]{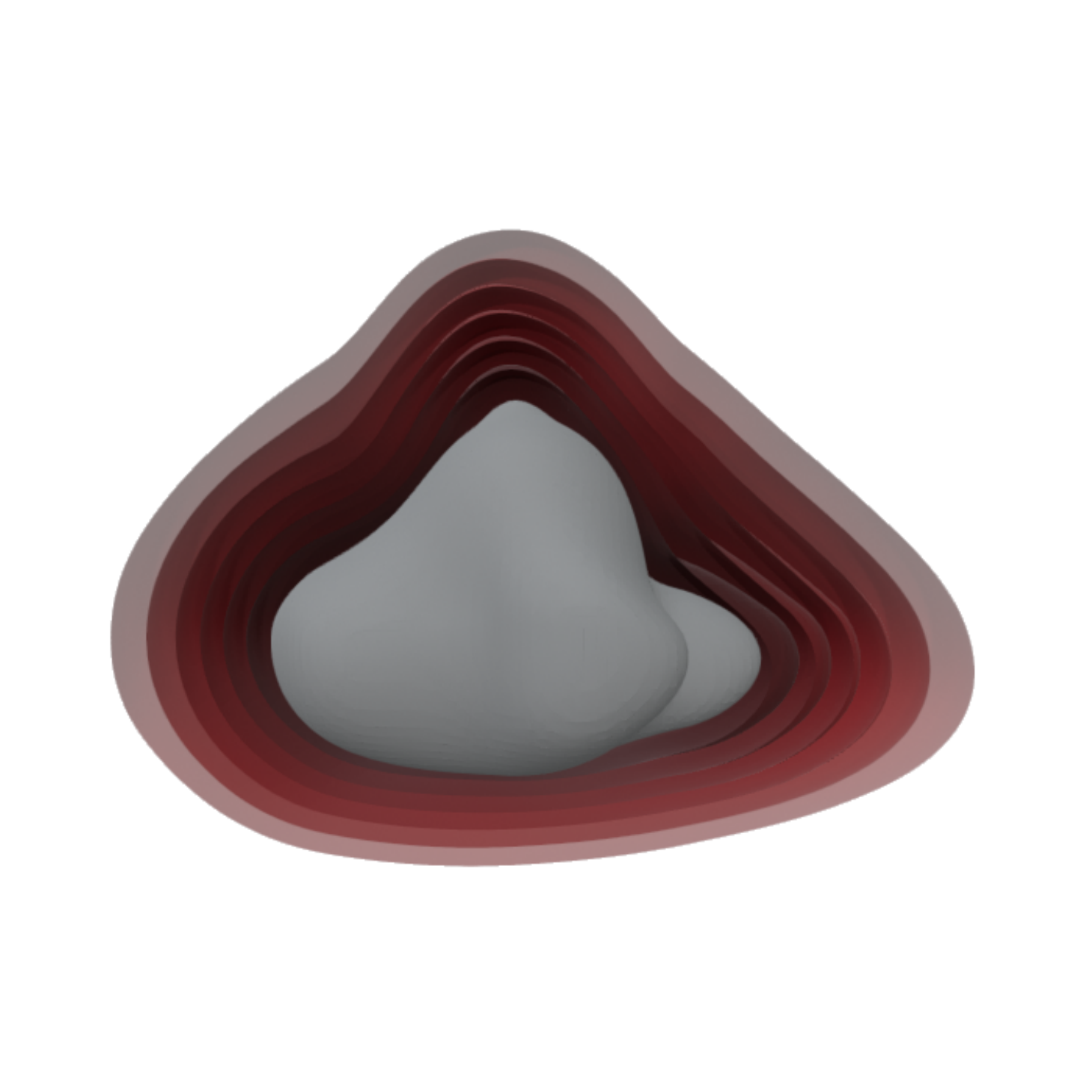}
        \caption{}
    \end{subfigure}
    \begin{subfigure}[t]{0.158\textwidth}
        \includegraphics[width=\textwidth]{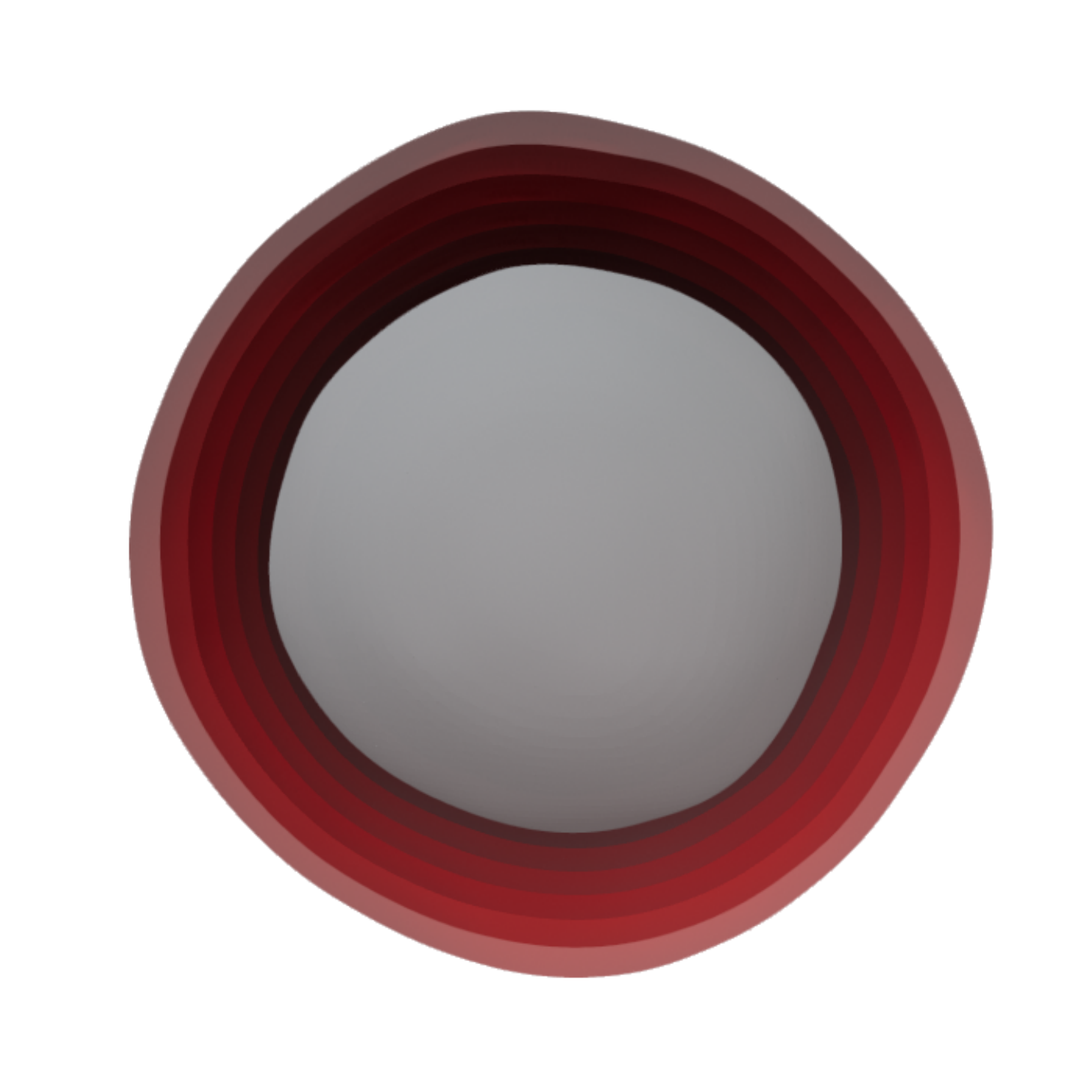}
        \caption{}
    \end{subfigure}
\caption{Mesh (gray) and level set (red) reconstruction results for objects from the YCB \cite{YCBds}
test set: (a) \texttt{003\_cracker\_box}, (b) \texttt{006\_mustard\_bottle}, (c) \texttt{011\_banana}, (d) \texttt{016\_pear}, (e) \texttt{019\_pitcher\_base}, (f) \texttt{021\_bleach\_cleanser}, (g) \texttt{048\_hammer}, (h) \texttt{035\_power\_drill}, (i) \texttt{063-a\_marbles}, (j) \texttt{053\_mini\_soccer\_ball}. Models were learned from $800$ non-uniformly sampled points and normals using 6 segments per input dimension.}
\label{fig:reconstruction}
\end{figure*}

We use quadratic error terms in order to evaluate the fitting of distance and normal data for $N$ incoming samples
\begin{align}
    c_{d}(\bm x_n) &= {\Big(\bm\Psi(\bm  x_n)\bm w\Big)^2},\\
    c_{g}(\bm x_n) &= \left|\left|\bm\nabla\bm\Psi(\bm x_n)\bm w - \bm g_n\right|\right|^2, 
\end{align}
with $\bm x_n = \left(x_n,y_n\right)$ denoting the $n$-th input sample, and $\bm g_n$ the corresponding sampled normal. An additional \textit{tension} term is used to constrain the curvature of the resulting distance field by minimizing the sum of second-order partial derivatives on $R$ control points
\begin{align}
    c_{t}(\bm x_r) &= {\left|\left|\bm{H}_{f}(\bm x_r)\right|\right|_F}^2,
\end{align}
$\forall r\in\{1,\ldots,R\}$, where $c_{t}(\bm x_r)$ represents the squared Frobenius norm of the corresponding Hessian matrix
\begin{align}
    \bm{H}_f(x, y) &= \left[\frac{\partial \bm{\nabla}f(x, y)}{\partial x}~ \frac{\partial \bm{\nabla}f(x, y)}{\partial y}\right].
\end{align}
The model weights can then be learned by minimizing a combined cost

\begin{align}
    \label{eq:cost}
     c = \lambda_d^2\sum_{n=1}^{N}c_{d}(\bm x_n)+\lambda_g^2\sum_{n=1}^{N}c_{g}(\bm x_n)+\lambda_t^2\sum_{r=1}^{R}c_{t}(\bm x_r),
\end{align}
with cost tuning coefficients $\lambda_d$, $\lambda_g$, and $\lambda_t$.
We construct our input vectors as concatenations of flattened distance, gradient, and tension features, denoted in italics
\begin{align}
    \bm{\Psi}^* &= \left[\it{\Psi} (\bm {x_1}) \cdots \it{\Psi}(\bm{x_N})\right]^\top,\\
    \bm{\nabla\Psi}^* &= \left[\it{\nabla\Psi}(\bm{x_1}) \cdots \it{\nabla\Psi}(\bm{x_N})\right]^\top,\\
    \bm{H_\Psi}^* &= \left[\it{H_\Psi}(\bm x_1) \cdots \it{H_\Psi}(\bm x_R)\right]^\top.
\end{align}
Finally, we can minimize \eqref{eq:cost} by calculating \eqref{eq:LSQ} using
\begin{align}
    \bm{A} = \begin{bmatrix} \lambda_d\bm{\Psi}^*\\ \lambda_g\bm{\nabla}\bm{\Psi}^*\\ \lambda_t\bm{{H}_\bm{\Psi}}^* \end{bmatrix},\quad
    \bm{s} = \begin{bmatrix} \bm{0}_{\bm\Psi} \\ \lambda_g\bm{g} \\ \bm{0}_{\bm H}\end{bmatrix},
\end{align}
where $\bm{g}$
is a vector of sampled normal components, and $\bm 0_{\bm\Psi}$ and $\bm 0_{\bm H}$ are zero vectors with lengths compatible with $\bm{\Psi}^*$ and $\bm{{H}_\bm{\Psi}}^*$, respectively.

\subsection{Incremental formulation}
Based on similar approaches used in the context of control \cite{Ting10b}, we update concatenated local models with an incremental variant of the least squares algorithm. Namely, we exploit the Sherman-Morrison-Woodbury relations \cite{Hager1989UpdatingTI} which connect subsequent inverses of a matrix after small-rank perturbations. This allows us to gradually refine an initial estimate by providing samples one by one or in batches.

After initializing the weight precision matrix $\bm P = \cov(\bm w)^{-1}$ to $\bm P_0$, it can be incrementally updated as
\begin{equation}
	\bm{P}_\new =
	\bm{P} - \underbrace{\bm{P}\bm{A}_n^\trsp \left(\sigma^{2}\bm{I}+\bm{A}_n\bm{P}\bm{A}_n^\trsp\right)^{-1}}_{\bm{K}} \bm{A}_n\bm{P},
\end{equation}
where $\sigma^2$ is the measurement noise variance, $\bm A_n$ the input matrix, and $\bm I$ an identity matrix of compatible dimensions. Starting from prior weights $\bm w = \bm w_0$, updates can then be calculated by using the Kalman gain $\bm K$
\begin{equation}
\bm{w}_\new = \bm{w} + \bm{K} \Big( \bm{s}_n - \bm{A}_n \bm{w} \Big).
\end{equation}
The above iterative computation has no requirement of storing the training points and enables us to impose priors on our model through $\bm P_0$ and $\bm w_0$. We impose a spherical prior on the weights and initialize the precision matrix as a scaled identity matrix
for recursive ridge regression.

\begin{algorithm}[h!]
\caption{Incremental computation of weights.}
\label{alg:splineRLS} 
$\bm{P} = \bm{P}_0$ \tcp{Initialize precision matrix}
$\bm{w} = \bm{w}_0$ \tcp{Initialize weights}
\For{$n \gets 1$ to $N$}{
	$\bm{A}_n = \bm{A}(\bm{x}_n)$ \tcp{Construct input matrix} 
	$\bm{K} = \bm{P}\bm{A}_n^\trsp \left(\sigma^2\bm{I}+\bm{A}_n\bm{P}\bm{A}_n^\trsp\right)^{-1}$\hspace{-2px}\tcp{Compute gain}
	$\bm{P} \leftarrow \bm{P} - \bm{K} \bm{A}_n \bm{P}$ \tcp{Update precision matrix}
	$\bm{w} \leftarrow \bm{w} + \bm{K} \Big( \bm{s}_n - \bm{A}_n \bm{w} \Big)$ \tcp{Update weights} 
}
\end{algorithm}

In order for our model to accurately approximate a distance function, the tension term needs to be enforced throughout the input space. We achieve this in an online setting by uniformly sampling a number of control points on the normal rays of incoming surface samples, as displayed in Figure \ref{fig:evolution}. The full computation steps are summarized in Algorithm \ref{alg:splineRLS}. We apply the same approach for three-dimensional inputs, with example reconstructions shown in Figures \ref{fig:bunny} and \ref{fig:reconstruction}.

\section{Evaluation} 

\subsection{Reconstruction accuracy}
We first evaluate the reconstruction accuracy of our approach by comparing it to two baseline methods used in online settings: 
\begin{enumerate}
    \item A neural network model based on \textit{iSDF} \cite{Ortiz:etal:iSDF2022}, using 4 hidden layers of 256 neurons, with corresponding positional embeddings and regularized loss.
    \item A GP model based on \textit{LogGPIS} \cite{wu2021LogGPIS} using the Matérn 3/2 kernel.
\end{enumerate}
Our comparison model utilizes cubic Bernstein polynomials with 6 segments per input dimension.
All models are implemented in \textit{PyTorch} and run on an \textit{NVIDIA GeForce MX550} GPU.

Accuracy evaluations are done on varying volumes of real point cloud data from the YCB dataset \cite{YCBds}.
The utilized test set consists of 10 household objects of diverse shapes, depicted in Figure \ref{fig:reconstruction}. All three methods are trained on the same depth-only point cloud and normal data, with ground truth SDFs reconstructed from high-definition meshes with 512k polygons. Note that \textit{LogGPIS} models unsigned Euclidean distance fields, and is therefore evaluated against absolute values of the ground truth.

\begin{figure}[h!]
\centering
    \includegraphics[width=0.76\linewidth]{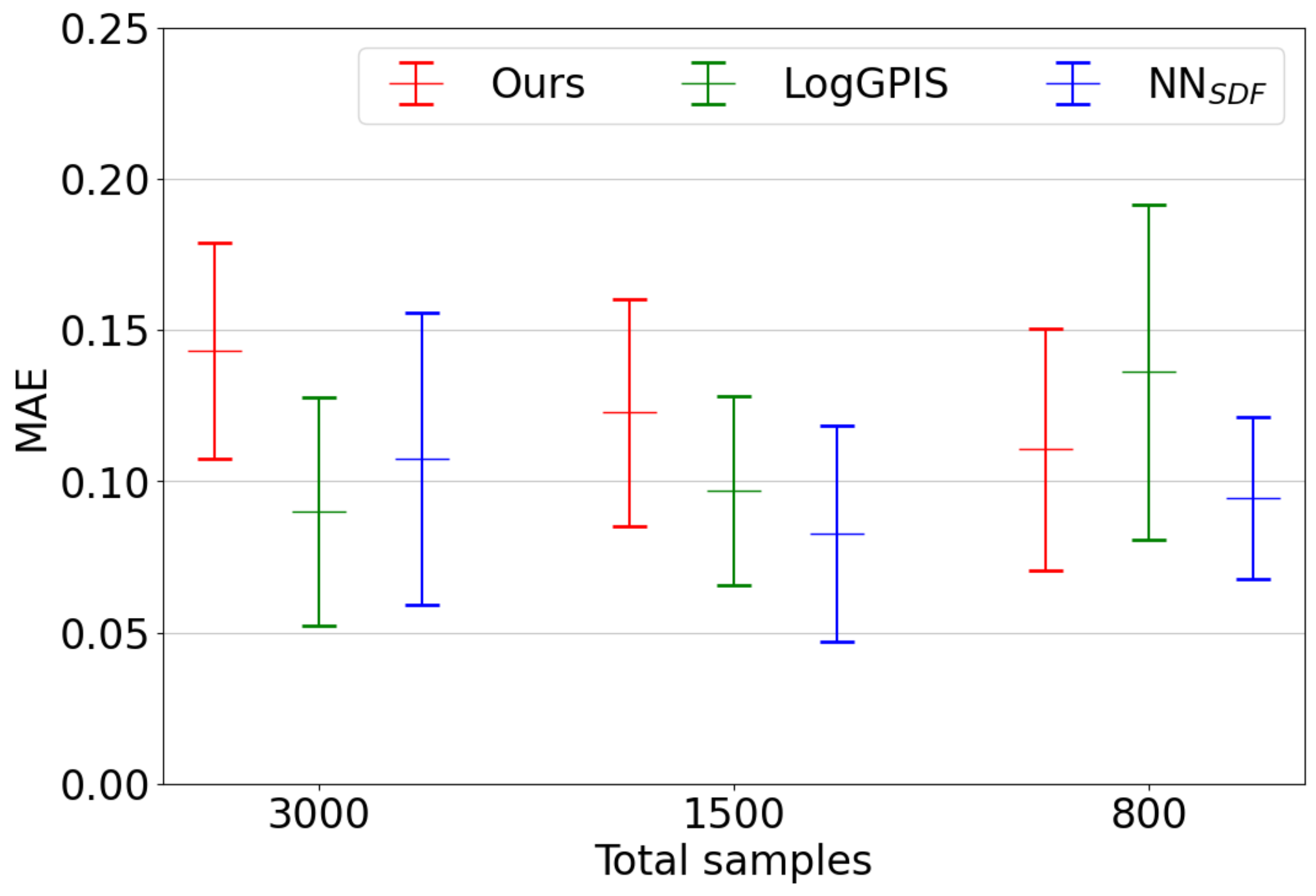}
\caption{
Distance reconstruction accuracy compared on varying amounts of training data.}
    \label{sub:MAEacc}
\end{figure}

To evaluate the reconstruction accuracy of our method we use error metrics similar to \cite{Ortiz:etal:iSDF2022}. Reconstructed distances are evaluated using the mean absolute error (MAE)
\begin{align}
    \text{MAE}(\bm x) = \big|\hat{s}(\bm x)-s(\bm x)\big|,
\end{align}
with $\hat{s}(\bm x)$ denoting the estimated signed distance at point $\bm x$, and $s(\bm x)$ the ground truth value. Figure \ref{sub:MAEacc} shows the resulting MAE comparisons. Qualitative reconstruction results of our method for all objects in the test set are shown in Figure \ref{fig:reconstruction}.

\noindent

\begin{figure}[h!]
    \centering
        \includegraphics[width=0.76\linewidth]{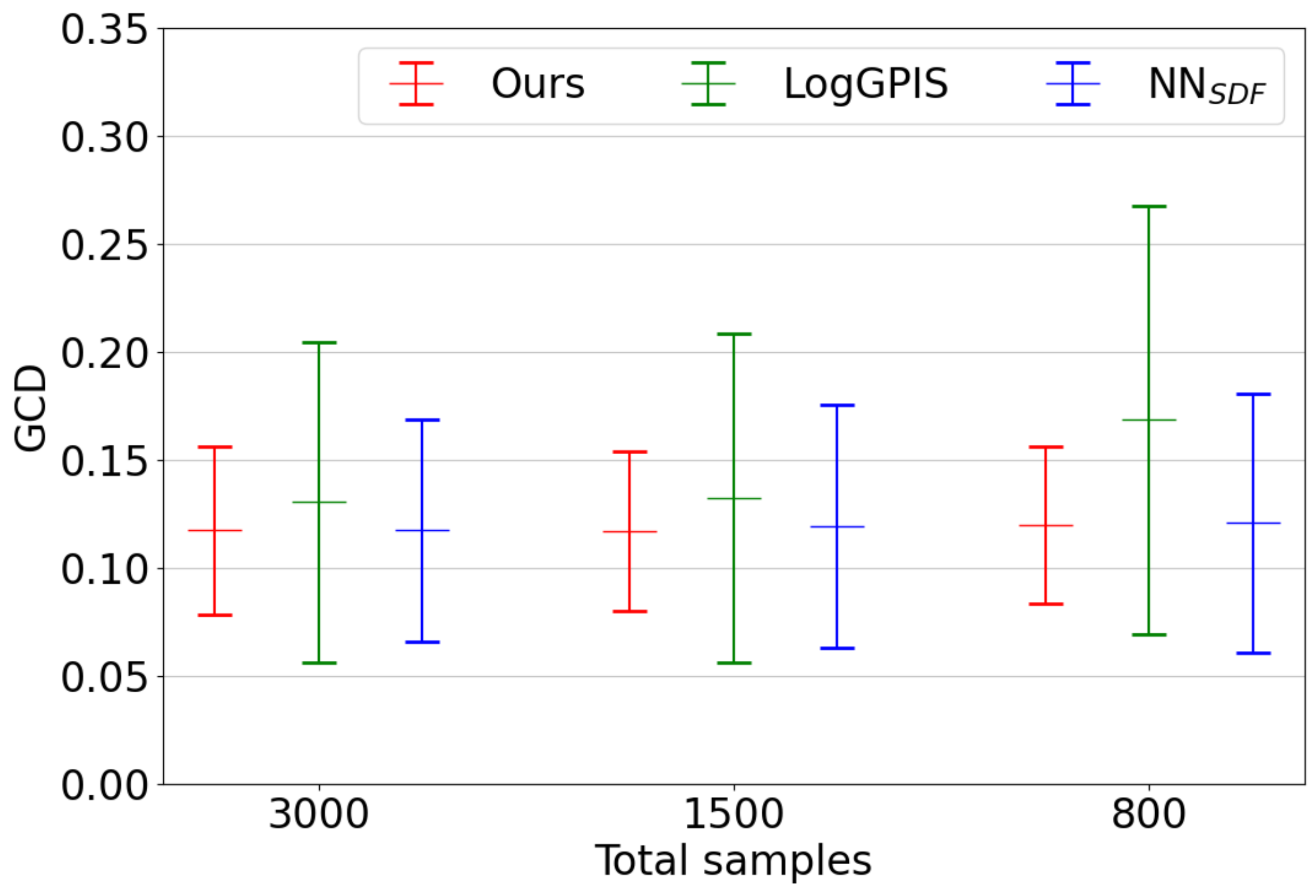}
    \caption{
Accuracy of reconstructed gradient fields with varying amounts of training data.}
        \label{sub:GCDacc}
\end{figure}

Additional comparisons are made with respect to reconstructed distance gradients by calculating the gradient cosine distance (GCD)
\begin{align}
    GCD(\bm x) = 1 - \frac{\nabla_{\bm x}\hat{s}(\bm x)\nabla_{\bm x}s(\bm x)}{\|\nabla_{\bm x}\hat{s}(\bm x)\|\|\nabla_{\bm x}s(\bm x)\|},
\end{align}
with $\nabla_{\bm x}\hat{s}(\bm x)$ denoting the estimated distance gradient, and $\nabla_{\bm x}s(\bm x)$ the corresponding ground truth. Figure \ref{sub:GCDacc} displays the GCD comparisons.

Many manipulation tasks inherently involve contact, thus requiring higher fidelity distance approximations closer to object surfaces. On the other hand, intermediary tasks such as reaching or collision avoidance often involve larger distances. Table \ref{tab:MAEnf} displays reconstruction accuracy near and far from object surfaces for different amounts of training samples.
\begin{table}[h!]
\centering
\vspace*{10px}
\begin{tabular}{|c|c|c|c|}
\multicolumn{4}{c}{$|s| < 0.05$} \\
\hline
Samples & Ours & GP & NN \\
\hline
3000 & $\num{0.0329} \pm \num{0.0201}$ & $\num{0.0620} \pm \num{0.0409}$ & $\num{0.0326} \pm \num{0.0152}$ \\
1500 & $\num{0.0332} \pm \num{0.0202}$ & $\num{0.0549} \pm \num{0.0405}$ & $\num{0.0326} \pm \num{0.0162}$ \\
800 & $\num{0.0338} \pm \num{0.0208}$ & $\num{0.0501} \pm \num{0.0369}$ & $\num{0.0330} \pm \num{0.0194}$ \\
\hline
\end{tabular}

\vspace{6px}
\begin{tabular}{|c|c|c|c|}
\multicolumn{4}{c}{$|s| > 0.05$} \\
\hline
Samples & Ours & GP & NN \\
\hline
3000 & $\num{0.1481} \pm \num{0.0337}$ & $\num{0.0916}\pm \num{0.0392}$ & $\num{0.1174} \pm \num{0.0561}$ \\
1500 & $\num{0.1261} \pm \num{0.0551}$ & $\num{0.1044} \pm \num{0.0325}$ & $\num{0.0818} \pm \num{0.0339}$ \\
800 & $\num{0.1079}\pm \num{0.0522}$ & $\num{0.1252} \pm \num{0.0461}$ & $\num{0.0798} \pm \num{0.0299}$ \\
\hline
\end{tabular}
\caption{Comparison of the mean absolute error (MAE) near and far from object surfaces for varying amounts of training samples.}
\label{tab:MAEnf}
\end{table}

\subsection{Computational requirements}
\label{sec:Comp}
Our representation relies on a sparse number of basis function weights as parameters, and does not store training data for prediction. The total number of parameters is further reduced by imposing constraints through concatenation, as described in Section \ref{sec:Method}.
For three-dimensional input and $S$ concatenated cubic polynomials, the number of utilized parameters corresponds to $N_w(S) = 8(S+1)^3$. The models utilized in our experiments therefore store and update only $N_w(6)=2744$ parameters. Comparatively, the $NN_{SDF}$ baseline has $200193$ learnable parameters, and \textit{LogGPIS} requires storing the training samples to inform mean and covariance predictions.

\begin{figure}[h!]
    \centering
    \includegraphics[width=0.8\linewidth]{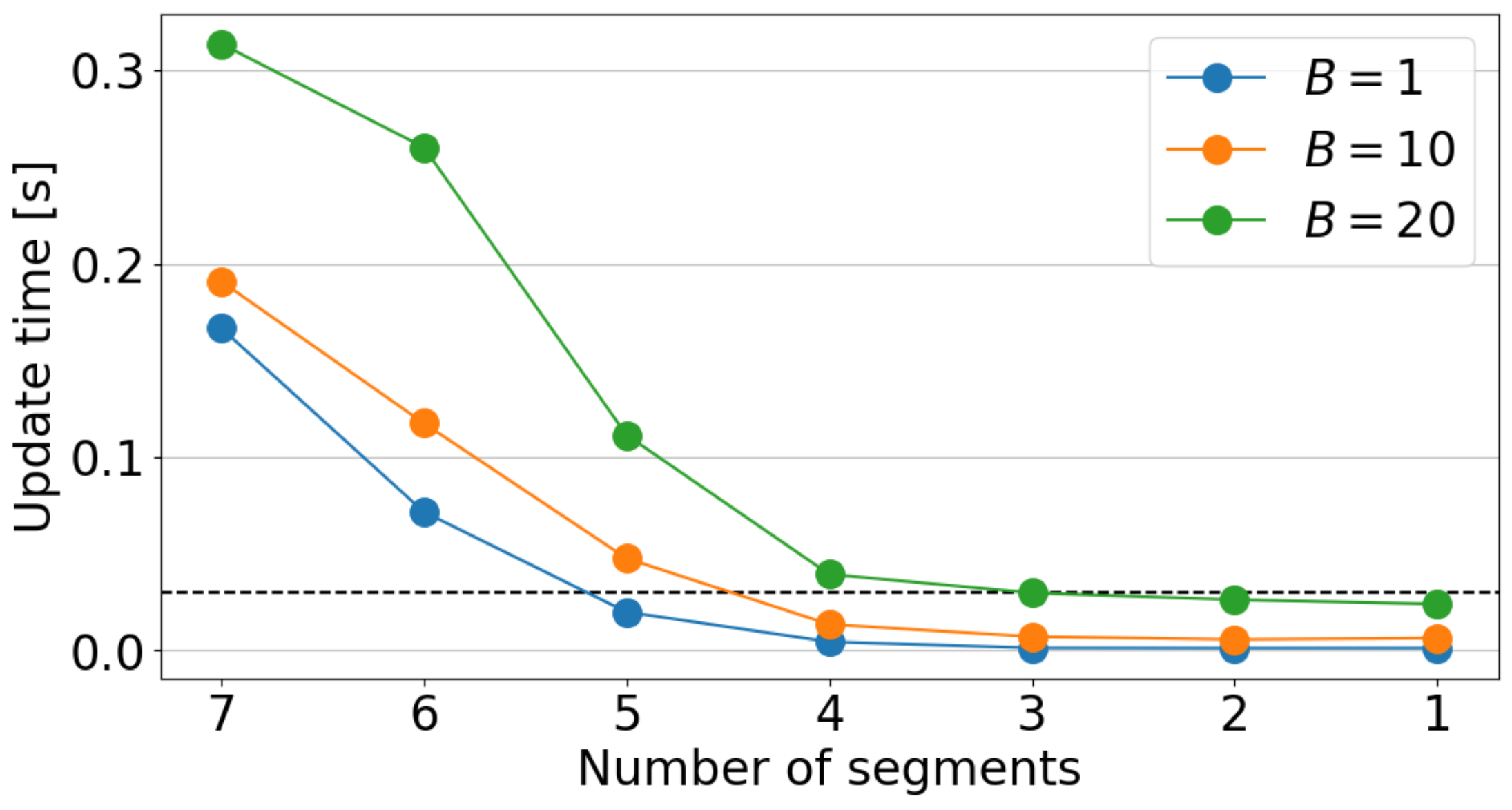}
    \caption{Update time for cubic polynomials with varying numbers of segments and different batch sizes. The desired real-time cutoff of $30~ms$ is denoted by a horizontal line.}
    \label{fig:segtimes}
\end{figure}

The computation time of updates to our model increases quadratically based on the number of weights. For a set input dimension and cubic polynomials, the time complexity of a single incremental update is $\mathcal O(B^3 + {N_w}^2B + {N_w} B^2)$, with
$B$ denoting the sample batch size.
Figure \ref{fig:segtimes} shows mean update times for different numbers of segments and varying batch sizes. Total training times of the evaluated models are compared in Table \ref{tab:training}.

\begin{table}[h!]
\centering
\begin{tabular}{|c|c|c|c|c|}
\multicolumn{5}{c}{Training time~[s]} \\
\hline
Samples & LogGPIS & NN$_{SDF}$ & Ours (S=6) & Ours (S=4) \\
\hline
3000 & $\num{28.58}$ & $\num{8.722}$ & $\num{32.36}$ & $\num{3.725}$\\
1500 & $\num{6.886}$ & $\num{7.002}$ & $\num{15.97}$ & $\num{1.835}$\\
800 & $\num{1.657}$ & $\num{4.720}$ & $\num{8.535}$ & $\num{0.9919}$\\
\hline
\end{tabular}
\caption{Total training time comparisons for varying numbers of training samples.}
\label{tab:training}
\end{table}
\begin{figure*}[h!]
    \centering
    \includegraphics[width=0.9\linewidth]{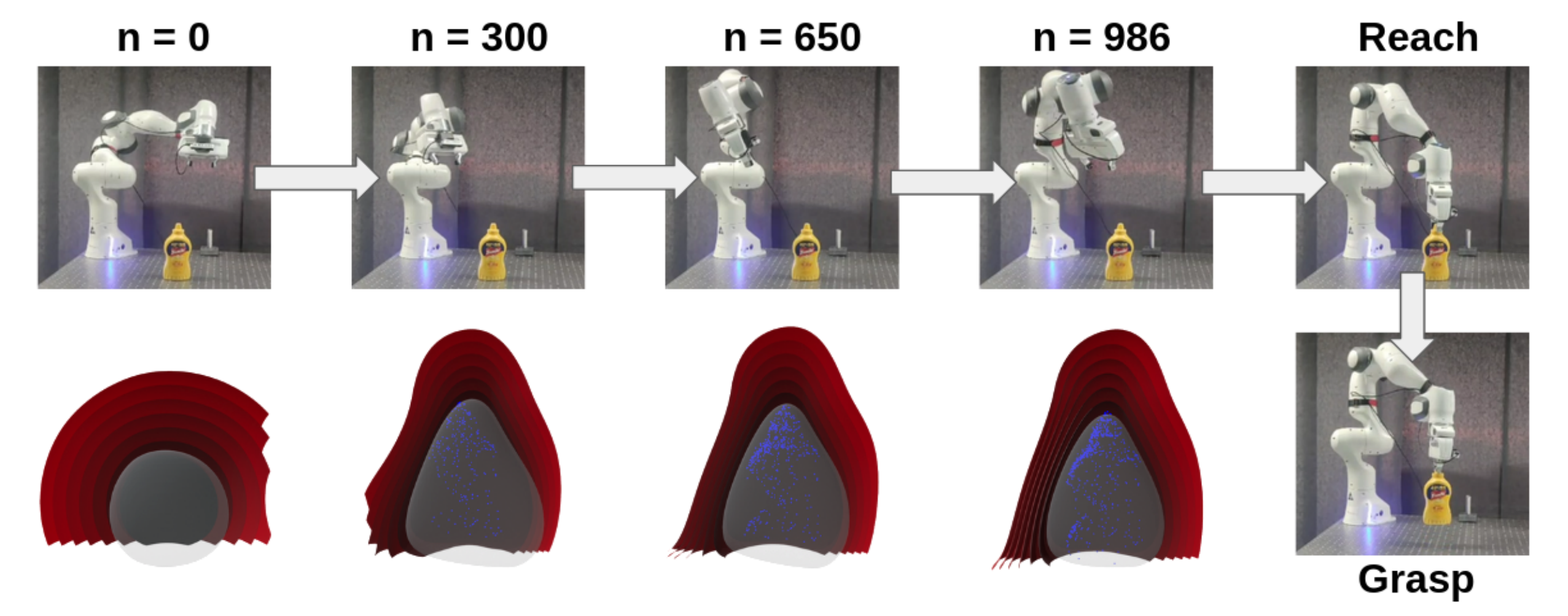}
    \caption{A visualization of our physical experiment and results, showing the evolution of a piecewise polynomial SDF model starting from a spherical prior. As $n$ samples are collected, the updated model is queried online to control the manipulator by following a trajectory tangential to the SDF level sets while keeping normal orientation. The reconstructed distance field is then used to generate and execute a collision-free grasp.
    }
    \label{fig:evolution_real}
\end{figure*}

For queries and reconstruction of distance and gradient fields, our method scales linearly with the number of weights, as we calculate only a weighted sum of basis functions. The time complexity of querying our model for a batch of $B$ points is $\mathcal O(N_wB)$, with mean time of $0.134$~ms for a single query of distance and gradient. Table \ref{tab:reconstruction} shows the mean time of full SDF reconstructions on a dense $128^3$ grid.
\begin{table}[h!]
\centering
\begin{tabular}{|c|c|c|c|c|}
\multicolumn{5}{c}{SDF reconstruction time~[s]} \\
\hline
Samples & LogGPIS & NN$_{SDF}$ & Ours (S=6) & Ours (S=4) \\
\hline
3000 & $\num{47.03}$ & $\num{0.9828}$ & $\num{5.7436}$ & $\num{5.4086}$\\
1500 & $\num{19.06}$ & $\num{0.9865}$ & $\num{5.9605}$ & $\num{5.5912}$ \\
800 & $\num{7.636}$ & $\num{0.9842}$ & $\num{5.8488}$ & $\num{5.3287}$\\
\hline
\end{tabular}
\caption{SDF reconstruction time with respect to varying numbers of training samples. 
}
\label{tab:reconstruction}
\end{table}
\subsection{Physical experiment}
\label{subsec:physical}
To test the viability of our method in a real-world setting, we perform a physical experiment in an online learning scenario using a Franka Emika 7-DoF manipulator. 
Starting from a spherical prior, we leverage the continuity and smoothness of our representation by querying gradients to execute a trajectory tangential to the SDF level sets. As the robot moves around the object of interest, partial-view depth data is collected by an in-hand Intel RealSense D415 sensor. The collected surface samples are used to incrementally update the SDF, directly shaping the trajectory of the manipulator.
Once the object shape is sufficiently explored, the reconstructed distance field is used to generate a valid grasp and execute it without colliding with the object.
To enable fast performance on CPU, we utilize cubic polynomials with 4 segments per input dimension and point-by-point updates.
The described learning and control framework is run with a mean timestep of $33.7$~ms on a $3.6$~GHz Intel Core i9-9900K CPU. Figure \ref{fig:evolution_real} shows our experiment setup and resulting SDF used to perform a grasp.

\section{Discussion}
\vspace{1px}
Figures \ref{sub:MAEacc} and \ref{sub:GCDacc} demonstrate that our incrementally learned representation can achieve similar distance and gradient reconstruction accuracy as baseline methods on the provided test set. Table \ref{tab:MAEnf} further shows that these results are consistent across data volumes, and that valid field reconstructions are maintained at different distances. 
As shown in Section \ref{sec:Comp}, our model relies on a low number of parameters to achieve these results, and the incremental learning scheme does not require storing the training data after model updates. Performance and accuracy of the proposed model can be balanced by adjusting the number of segments or degree of utilized polynomial basis functions. Inference time scales linearly with the number of weights and batch size, allowing for fast queries that correspond to calculating a weighted sum of basis functions. This is further reflected by full SDF reconstruction times in Table \ref{tab:reconstruction}. At the expense of increased memory usage, faster large-scale inference (e.g., for visualization purposes) can be achieved by precomputing the basis function values. Additional efforts in performance optimization might consider leveraging local weight updates and further parallelization.

Qualitative reconstructions in Figure \ref{fig:reconstruction} display visually accurate distance and mesh reconstructions across the test set, with a noted over-smoothing effect due to polynomial approximation on a low-resolution grid. The over-smoothing effect becomes less pronounced as spatial resolution is increased by further segmenting the input space. However, as our implementation relies on a uniform grid for segmenting the input space, this can rapidly increase the number of weights and result in higher computation times reflected by Figure \ref{fig:segtimes} and training time comparisons in Table \ref{tab:training}. Scaling the model for higher accuracy and larger or more complex environments might therefore require combining the online paradigm with adaptive grids or hierarchical models such as octrees, which will be a topic of our future work.

The physical experiment showcases the interplay of incremental updates and fast queries of analytical gradients to adapt the trajectory of a manipulator as surface samples are collected. Furthermore, it validates the use of an incrementally updated prior for motion planning with noisy and partial point cloud data. For fast performance on CPU, the utilized model was reduced to 4 segments per input dimension, showing that the resulting distance and gradient fields are still accurate enough for a surveying and grasping task. This demonstrates the viability of using a basis function representation of the SDF in a simple manipulation scenario, and opens the way toward methods further leveraging the properties of Bernstein polynomials and related families of basis functions for learning and reconstruction. Future work will focus on additional quantitative evaluation to investigate how accuracy and robustness of the proposed method scale to more complex manipulation tasks, higher levels of measurement noise, and more dynamic scenes. Lastly, we intend to explore contact-rich scenarios where incremental updates with tactile and proximity data might be of particular interest, and investigate the use of more informative priors conditioned on modalities such as RGB or tactile data.

\section{Conclusion}
This paper presented an online formulation of signed distance fields using piecewise polynomial basis functions. Starting from an arbitrary prior shape, our method incrementally builds a smooth distance representation from incoming surface points. It offers analytical access to gradients and ensures a desired order of continuity through constraints on the basis function weights. Furthermore, performance of the underlying model can be balanced through interpretable hyperparameters. Our results show that a low number of parameters can be used to achieve similar reconstruction accuracy to Gaussian process and neural network baselines on a test set of household objects.
Finally, we demonstrated the use of the online basis function representation in a physical surveying and grasping task with noisy partial observations, and discussed possible extensions for further scalability.

\bibliographystyle{ieeetr}

\bibliography{root}

\end{document}